# Improved Onlooker Bee Phase in Artificial Bee Colony Algorithm

Sandeep Kumar
Jagannath University
Chaksu, Jaipur, Rajasthan,
India - 303901

Vivek Kumar Sharma, Ph.D
Jagannath University
Chaksu, Jaipur, Rajasthan,
India - 303901

Rajani Kumari
Jagannath University
Chaksu, Jaipur, Rajasthan,
India - 303901

## ABSTRACT
Artificial Bee Colony (ABC) is a distinguished optimization strategy that can resolve nonlinear and multifaceted problems. It is comparatively a straightforward and modern population based probabilistic approach for comprehensive optimization. In the vein of the other population based algorithms, ABC is moreover computationally classy due to its slow nature of search procedure. The solution exploration equation of ABC is extensively influenced by a arbitrary quantity which helps in exploration at the cost of exploitation of the better search space. In the solution exploration equation of ABC due to the outsized step size the chance of skipping the factual solution is high. Therefore, here this paper improve onlooker bee phase with help of a local search strategy inspired by memetic algorithm to balance the diversity and convergence capability of the ABC. The proposed algorithm is named as Improved Onlooker Bee Phase in ABC (IoABC). It is tested over 12 well known un-biased test problems of diverse complexities and two engineering optimization problems; results show that the anticipated algorithm go one better than the basic ABC and its recent deviations in a good number of the experiments.

## General Terms
Computer Science, Nature Inspired Algorithms, Meta-heuristics

## Keywords
Artificial bee colony algorithm, Swarm intelligence, Evolutionary computation, Memetic algorithm

## 1. INTRODUCTION
Swarm Intelligence has turned out to be a promising and out of the ordinary area in the field of nature inspired techniques that are used to get to the bottom of optimization problems at some stage in the past decade. It is motivated by the communal behavior of social living things. Swarm based optimization algorithms hit upon solution by collaborative trial and error. Social creatures employ their talent of social learning to get to the bottom of complex tasks. Peer to peer learning conduct of social colonies is the most important driving force behind the expansion of many efficient swarm based optimization algorithms. Investigators have investigated such behaviors and proposed algorithms that can be used to get to the bottom of nonlinear, non-convex or discrete optimization problems. Prior research [1, 2, 3, 4] have revealed that algorithms based on swarm intelligence have enormous prospective to find solutions of factual world optimization problems. The algorithms that have materialized in recent years comprise ant colony optimization (ACO) [1], particle swarm optimization (PSO) [2], bacterial foraging optimization (BFO) [5] etc.

Artificial bee colony (ABC) optimization algorithm initiated by Karaboga [6] is a ground-breaking addition in this group. This algorithm is stimulated by the conduct of honey bees while searching a quality food source. Similar to other inhabitants based optimization strategy, ABC consists of a population of impending solutions. The budding solutions are food sources of honey bee insects. The fitness is established in terms of the eminence (nectar amount) of the food source. ABC is somewhat a straightforward, fast and population based stochastic investigation technique in the pasture of nature inspired algorithms. There are two elementary course of action which compels the swarm to keep informed in ABC: the deviation process, which enables exploring poles apart areas of the search space, and the assortment process, which make certain the exploitation of the previous understanding. However, it has been shown that the ABC may occasionally stop proceeding toward the global optimum even though the population has not congregated to a local optimum [7]. It can be observed that the solution search process of original ABC algorithm is fine at exploration but pitiable at exploitation [8].

For that reason, to maintain the appropriate steadiness between exploration and exploitation behavior of ABC, it is highly required to develop a local search approach in the fundamental ABC to exploit the search region. In earlier period, very few endeavors have been done on this trend. Kang et al. [9] anticipated a Hooke Jeeves Artificial Bee Colony algorithm (HJABC) for numerical optimization. HJABC integrates a new local search modus operandi which is based on Hooke Jeeves method (HJ) [10] with the basic ABC. Further, J. C. Bansal et al. [34] introduced an alternate of the original ABC named Memetic search in Artificial Bee Colony algorithm. In this work, the authors incorporated memetic search procedure inspired by Golden Section Search (GSS) process [36]. S. Kumar et al. proposed a novel hybrid of crossover based ABC [11] for global optimization by incorporating crossover phase from genetic algorithm.

In this paper, new search strategy is proposed. The proposed local search strategy is used in place of onlooker bee phase. Further, the proposed algorithm is compared by experimenting on 14 un-biased test problems (i.e. the problems which solutions do not exist at starting point, axes or diagonal) to the basic ABC and its recent variants named, Memetic ABC (MeABC) [34], Randomized Memetic ABC (RMABC) [13], Modified ABC (MABC) [14] and Fitness Based Position Update in ABC (FPABC) [15].

Rest of the paper is systematized as follows: Sect. 2 describes brief overview of the basic ABC. Memetic algorithms explained in Sect. 3. Improved Onlooker phase in ABC (IoABC) is proposed and tested in Sect. 4. In Sect. 5, a comprehensive set of experimental results are provided. Finally, in Sect. 6, paper is concluded.

## 2. ARTIFICIAL BEE COLONY ALGORITHM

The ABC algorithm is relatively up to date swarm intelligence based algorithm. The algorithm is stimulated by the intellectual food foraging behavior of honey bees. In ABC, each solution of the problem is known as food source of honey bee insects. The fitness is established in terms of the quality of the food source. In ABC, honey bees are categorized into three classes namely employed bees, onlooker bees and scout bees. The counts of employed bees are equal to the onlooker bees. The employed bees are the bees which search the food source and congregate the information about the eminence of the food source. Onlooker bees are the bees which reside in the beehive and search the food sources on the basis of the information collected by the employed bees. The scout bee searches new food sources haphazardly in places of the discarded foods sources. Analogous to the other population-based algorithms, ABC solution search process is a step by step iterative process. Following, initialization of the ABC parameters and swarm, it requires the monotonous iterations of the three phases namely employed bee phase, onlooker bee phase and scout bee phase.

### 2.1 Correspondence of Artificial Bee Colony Algorithm and Honey Bee

The innovative model proposed by D. Karaboga [6] consists of three major rudiments: employed and unemployed foragers, and food sources. The employed bees are collaborator with an appropriate food source. Employed bees have intimate understanding about food source. Utilization of food sources done by employed bees. When a food source discarded employed bee become unemployed. The unemployed foragers are bees having no information on the subject of food sources and searching for a food source to take advantage of it. We can categorize unemployed bees in two categories: scout bees and onlooker bees. Scout bees search at hit and miss for new food sources adjoining the hive. Onlooker bees examine the waggle dance in beehive, to select a food source for exploitation. The third element is the prosperous food sources in propinquity to their hive. Relatively in the optimization framework, the number of food sources (that is either employed or onlooker bees) in ABC algorithm, is corresponding to the number of solutions in the population. Additionally, the locality of a food source corresponds to the position of a complimentary solution to the optimization problem, in view of the fact that the trait of nectar of a food source correspond to the fitness cost (quality) of the allied solution.

### 2.2 Phases of Artificial Bee Colony Algorithm

The investigation course of action of ABC follows three most important steps [6]:

- Send the employed bees to a food source to collect information about it and decide the nectar quality;
- Onlooker bees decide on the food sources subsequent to gathering information from employed bees and deciding the nectar eminence;
- Find out the scout bees and make use of them onto achievable food sources.

The locality of the food sources are capriciously selected by the bees at the preliminary stage and their nectar qualities are measured. The employed bees then share the nectar information of the sources with the onlooker bees waiting at the dance vicinity within the beehive. After distribution of this information, each employed bee returns to the food source checked at some stage in the previous cycle, as the location of the food source had been recalled and at that time selects new food source using its observed information in the neighborhood of the present food source. At the last stage, an onlooker bee uses the information retrieved from the employed bees at the dance area to select a good food source. The opportunity for the food sources to be nominated rose with rise in its quality of nectar. Hence, the employed bee with information of a food source with the peak quality of nectar employs the onlookers to that food source. It subsequently chooses another food source in proximity of the one presently in her memory depending on experiential information. A new food source is erratically generated by a scout bee to swap the one deserted by the onlooker bees.

#### 2.2.1 Initialization of Swarm

The ABC algorithm has three major parameters: the number of food sources (population), the quantity of test subsequent to which a food source is treated to be deserted (limit) and the termination criteria (maximum number of cycle). In the original ABC proposed by D. Karaboga [6], the number of food sources is equal to the employed bees or onlooker bees. In the beginning it think about an regularly dealt swarm of food sources (SN), where every food source $x_i$ (i = 1, 2 ...SN) is a D-dimensional vector. Each food source is generated using subsequent equation [7]:

$$x_{ij} = x_{minj} + rand[0,1](x_{maxj} - x_{minj}) \qquad (1)$$

Here rand[0,1] is a function that gives an equally distributed random number in range [0,1].

#### 2.2.2 Employed Bee

Employed bees phase bring up to date the present solution based on the information of individual understandings and the fitness value of the newly found solution. New food source with higher fitness value replace the existing one. The situation update equation for $j^{th}$ dimension of $i^{th}$ candidate during this phase is shown below [7]:

$$V_{ij} = x_{ij} + \varphi_{ij}(x_{ij} - x_{kj}) \qquad (2)$$

Where $\varphi_{ij}(x_{ij} - x_{kj})$ is known as size of step, k ∈ {1, 2, ..., SN}, j ∈ {1, 2, ...,D} are two haphazardly preferred indices. k ≠i make sure that step size has some problem-solving upgrading.

#### 2.2.3 Onlooker Bee

The number of food sources for onlooker bee is alike as the employed bee. For the duration of this phase all employed bee share fitness information of new food sources with onlooker bees. Onlooker bees determine the selection probability of every food source engendered by the employed bee. The superlative food source is elected by the onlooker. There are number of scheme for computation of probability, but it must include fitness. Probability of each food source is decided using its fitness as follow [7]:

$$P_i = \frac{fit_i}{\sum_{i=1}^{SN} fit_i} \qquad (3)$$

#### 2.2.4 Scout Bee Phase

If the locality of a food source is not updated for a predefined number of iterations, then the food source is considered as deserted and scout bees phase is initialized. During this phase the bee coupled with the deserted food source transformed into scout bee and the food source is replaced by the capriciously chosen food source inside the search space. In ABC, the predefined number of cycles is an important control parameter which is called limit for rejection. Now the scout bees substitute the deserted food source with new one using following equation [7].

$$x_{ij} = x_{minj} + rand[0,1](x_{maxj} - x_{minj}) \forall j = 1,2,.D \quad (4)$$

As per the above discussion, it is unambiguous that in ABC search process there are three most important control parameters: the number of food sources SN (same as number of onlooker or employed bees), the limit and the maximum number of cycles. The major steps of the ABC algorithm are outlined as follow [7]:

---

**Algorithm 1:** Artificial Bee Colony Algorithm

Initialize all parameters;
Repeat while Termination criteria is not meet
    Step 1: Employed bee phase for computing new food sources.
    Step 2: Onlooker bees phase for updating location the food sources based on their amount of nectar.
    Step 3: Scout bee phase for searching new food sources in place of rejected food sources.
    Step 4: Memorize the best food source identified so far.
End of while
Output: The best solution identified so far.

## 3. MEMETIC ALGORITHMS

Memetic Algorithms (MAs) was the name introduced by P.A. Moscato [18] to a class of stochastic global search techniques that, commonly speaking, coalesce within the skeleton of Evolutionary Algorithms (EAs) the profit of problem-oriented local search heuristics and multi-agent systems. In ethnic improvement processes, information is processed and extravagant by the communicating parts; it is not only transmitted impassive between entities. This enhancement is recognized in MAs by taking on board heuristics, approximation algorithms, metaheuristics, local search techniques, particular recombination operators and truncated exact search methods. In aspect, more or less all MAs can be illustrated as a search technique in which a population of optimizing operator work together and compete. MAs have been successfully imposed to a large scale domains that encircle problems in combinatorial optimization, like E. Burke, J. Newall, R. Weare [19] used memetic algorithm for university exam timetabling. R. Cheng, M. Gen [21] use memetic algorithms for scheduling parallel machine. R. Carr, W. Hart, N. Krasnogor, E. Burke, J. Hirst, J. Smith [20] applied a memetic evolutionary algorithm for alignment of protein structures. C. Fleurent, J. Ferland [22] developed a hybrid of Genetic algorithm with graph coloring algorithm. It is also applied for solving travelling salesman problem [23], back routing in telecommunication [24], bin packing [25], VLSI floor planning [26] continuous optimization [27],[28], dynamic optimization [29] and multi-objective optimization [30].

Exploitation capability of evolutionary computing enhanced at large scale in collaboration to memetic algorithm. Application area of memetic algorithms is endlessly expending after its initiation. Y. Wang et al [31] developed a memetic algorithm for the maximum diversity problem based on Tabu search. X. Xue et al [32] Optimize ontology alignment with help of Memetic Algorithm based on partial reference alignment. O. Chertov and Dan Tavrov [33] anticipated a memetic algorithm for solution of the task of providing group anonymity. J. C. Bansal et al [34] included Memetic search in artificial bee colony algorithm. F. Kang et al [9] anticipated a new memetic algorithm HJABC inspired by hooke-jeeves method. I. Fister et al [35] anticipated a memetic ABC algorithm for large-scale global optimization.

Memetic algorithm proposed by Kang et al. [9] incorporates Hooke Jeeves [10] local search method in Artificial Bee Colony algorithm. HJABC is a hybrid algorithm of intensification search based on the Hooke Jeeves pattern search and the ABC. HJABC modify the fitness (Fiti) calculation function incorporate the Hooke-Jeeves local search in original ABC. HJABC contains combination of exploratory move and pattern move to search optimum result of problem. The first, step exploratory move think about one variable at a time in order to choose appropriate direction of search process. The second step is pattern search to speed up search in decisive direction by exploratory move. These two steps repeated until the rejection criteria meet. The Hooke-Jeeves pattern move is a contentious attempt of the algorithms for the exploitation of promising search directions as it collect information from previous successful search iteration.

The memetic search in ABC (MeABC) anticipated by J. C. Bansal et al [34] stimulated by Golden Section Search (GSS) process [36]. In MeABC only the superlative particle of the current swarm updates itself in its immediacy. MeABC also modify position update equation as per the following equation in order to control step size.

$$x'_{ij} = x_{ij} + \phi_{ij}(x_{ij} - x_{kj}) + \psi_{ij}(x_{bestj} - x_{ij}) \quad (5)$$

Here $\Psi_{ij}$ is an arbitrary number in interval [0, C], for some positive constant C.

## 4. IMPROVED ONLOOKER BEE PHASE IN ARTIFICIAL BEE COLONY (IoABC) ALGORITHM

Exploration and exploitation are the two vital characteristics of the population-based optimization algorithms such as GA (16), PSO (2), DE (17), and BFO (5). In these optimization algorithms, the exploration refers to the ability to investigate the various unknown regions in the solution space to discover the global optimum. The exploitation is the ability to use the knowledge of the previous good solutions to find better solutions and exploration is the process that spread the search space. The exploration and exploitation are two contradictory concepts, and in order to accomplish better optimization performance, these two abilities must remain in balance. Dervis Karaboga and Bahriye Akay (7) experienced different modifications of ABC for global optimization and bring into being that the ABC shows poor performance and remains inefficient during the exploration of the search space. In ABC, any probable solution updates itself using the information provided by a randomly selected probable solution within the in progress swarm. In this development, a step size which is a linear combination of a casual number $\phi_{ij} \in [-1, 1]$, current solution and a haphazardly selected solution are used. Now the quality of the modified solution generally depends upon this step size. If the step size is too large then updated solution can surpass the true solution. Large step size may takes place if the difference of current solution and randomly selected solution is large with high unqualified value of $\phi_{ij}$, and if this step size is too small then the convergence rate of ABC may considerably decrease as it takes more time to move towards optimum value. An appropriate sense of balance in this step size can balance the exploration and exploitation capability of the ABC at the same time. But, since this step size consists of random element so the balance cannot be done by hand. The exploitation capability can be improved by incorporation of a local search algorithm with the ABC algorithm. For that reason, this paper introduce, an improved onlooker bee phase inspired by modified GSS process to balance the diversity and convergence speed of ABC. It modifies the range of two parameters in GSS process and applies GSS based search process in onlooker bee phase. For that reason, in these modifications, better solutions get more chance in search process and minimize the threat of less stability here. The improved search strategy in ABC is outlined as follow:

Step 1: Initialize the population of N evenly disseminated individuals. Each individual $x_{ij}$ is a food source (i.e. required solution) and has D number of attributes. D is identified as the dimension of the problem. $i^{th}$ solution in $j^{th}$ dimension denoted as $x_{ij}$. Where $j \in \{1, 2, \ldots, D\}$

$$x_{ij} = x_{minj} + rand[0,1](x_{maxj} - x_{minj})$$

Step 2: Calculate approximately the fitness of each and every individual solution using the following method,

if (soln_value >= 0)

then

$$fit_i = \emptyset \times \left(\frac{1}{2 \times soln_{val} + 1}\right) + (1 - \emptyset) \times (1 + fabs\left(\frac{1}{soln_{val}}\right))$$

else

$$fit_i = (1 - \emptyset) \times \left(\frac{1}{2 \times soln_{val} + 1}\right) + \emptyset \times (1 + fabs\left(\frac{1}{soln_{val}}\right))$$

Step 3: Each employed bee, placed at a food source that is different from others, search in the proximity of its current position to find a better food source. For each employed bee, generate a new solution, $v_{ij}$ around its current position, $x_{ij}$ using the following formula.

$$v_{ij} = x_{ij} + \emptyset_{ij}(x_{ij} - x_{kj})$$

Here, $k \in \{1, 2, \ldots, N\}$ and $j \in \{1, 2, \ldots, D\}$ are randomly chosen indices. N is number of employed bees. $\phi_{ij}$ is a uniform arbitrary number from [-1, 1].

Step 4: Compute the fitness of both $x_{ij}$ and $v_{ij}$. Apply greedy selection strategy to select better one of them.

Step 5: Calculate and normalize the probability values, $P_{ij}$ for each solution $x_i$ using the following formula.

$$P_{ij} = \emptyset \times \left(\frac{fit_i}{Max_{fitness}}\right) + (1 - \emptyset) \times (fit_i / \sum_{i=1}^{N} fit_i)$$

Here $\phi$ is a random number in range [0,1]

Step 6: Assign each onlooker bee to a solution, $x_i$ at random with probability proportional to $P_{ij}$. Apply improved search phase inspired by GSS process. Take a=-1.2, b 1.2 and $\phi$ = rand[0.55,0.65]. Compute $f_1$=b-((b-a)*$\phi$) and $f_2$=a+ ((b-a)*$\phi$).

Repeat while termination criteria meet

Calculate value of function based on $f_1$ and $f_2$.

If $f_{1.val} < f_{2.val}$ then

    b = $f_2$ and the solution lies in the range [a, b]

else

    a = $f_1$ and the solution lies in the range [a, b]

Modify the position of solution using following equation

$$x'_{ij} = x_{ij} + (x_{ij} - x_{kj}) \times f_l$$

Here k=rand[0,1]*Food Number, j=rand[0,1]*dimension and l={1,2}.

Step 7: Arrange new food sources, $v_{ij}$ for each onlooker bee.

Step 8: Compute the fitness of each onlooker bee, $x_{ij}$ and the new solution, $v_{ij}$. Select the fittest one using greedy selection process.

Step 9: If a particular solution $x_{ij}$ has not been improved over a predefined number of cycles, then select it for rejection. Replace the solution by placing a scout bee at a food source generated evenly at random within the search space using

$$x_{ij} = x_{minj} + rand[0,1](x_{maxj} - x_{minj})$$

for j = 1, 2,……,D

Step 10: Keep track of the best food sources (solution) found so far.

Step 11: Check termination criteria. If the best solution found is acceptable or reached the maximum iterations, stop and return the best solution found so far. Otherwise go back to step 2 and repeat again.

# 5. EXPERIMENTAL RESULTS
## 5.1 Test problems under consideration

Artificial Bee Colony algorithm with improvement in onlooker bee phase applied to the twelve benchmark functions for whether it gives better result or not at different probability and also applied for two real world problems. Benchmark functions taken in this paper are of different characteristics like uni-model or multi-model and separable or non-separable and of different dimensions. In order to analyze the performance of IoABC, it is applied to global optimization problems ($f_1$ to $f_{14}$) listed in Table 1. Test problems $f_1 - f_{14}$ are taken from [37-39].

**Compression Spring ($f_{13}$):** The compression spring problem [37] minimizes the weight of a compression spring that is subjected to constraints of shear stress, surge frequency, minimum deflection and limits on outside diameter and on design variables. In case of compression spring three design variables considered: The diameter of wire($x_1$), mean coil diameter ($x_2$) and count of active coils ($x_3$). Simple mathematical representation of this problem is:

$$x_1 \in \{1, 2, 3, \ldots, 70\} granularity\ 1,$$

$$x_2 \in [0.6; 3], x_3 \in [0.207; 0.5] granularity\ 0.001$$

and four constraints

$$g_1 = \frac{8c_f F_{max} x_2}{\pi x_3^3} - S \leq 0, g_2 = l_f - l_{max} \leq 0$$

$$g_3 = \sigma_p - \sigma_{pm} \leq 0, g_4 = \sigma_w - \frac{F_{max} - F_p}{K} \leq 0$$

Where $c_f = 1 + \frac{0.75 \times (x_3)}{x_2 - x_3} + 0.615 \times \frac{x_3}{x_2}, F_{max} = 1000,$

$$S = 189000, l_f = \frac{F_{max}}{K} + 1.05 \times (x_1 + 2) \times x_3, l_{max} = 14,$$

$$\sigma_p = \frac{F_p}{K}, \sigma_{pm} = 6, F_p = 300, K = 11.5 \times 10^6 \times \frac{x_3^4}{8x_1 x_2^2}, \sigma_w = F_p/K$$

And the function to be minimized is

$$f_{13}(X) = \pi^2 \times (x_2 x_3^2 (x_1 + 2)/4$$

The best ever identified solution is (7, 1.386599591, 0.292), which gives the fitness value $f = 2.6254$ and $1.0E\text{-}04$ is tolerable error for compression spring problem.

**Pressure Vessel Design ($f_{14}$):** The problem of minimizing total cost of the material, forming and welding of a cylindrical vessel [37]. In case of pressure vessel design generally four design variables are considered: shell thickness ($x_1$), spherical head thickness ($x_2$), radius of cylindrical shell ($x_3$) and shell length ($x_4$). Simple mathematical representation of this problem is as follow:

$$f_9 = 0.6224 x_1 x_3 x_4 + 1.7781 x_2 x_3^2 + 3.1611 x_1^2 x_4 + 19.84 x_1^2 x_3$$

Subject to

$$g_1(x) = 0.0193 x_3 - x_1, g_2(x) = 0.00954 x_3 - x_2,$$

$$g_3(x) = 750 * 1728 - \pi x_3^2 (x_4 + \frac{4}{3} x_3)$$

**Table 1. Test Problems**

| Test Problem | Objective Function | Search Range | Optimum Value | D | Acceptable Error |
|---|---|---|---|---|---|
| Zakharov | $f_1(x) = \sum_{i=1}^{D} x_i^2 + (\sum_{i=1}^{D} \frac{i x_i}{2})^2 + (\sum_{i=1}^{D} \frac{i x_1}{2})^4$ | [-5.12, 5.12] | $f(0) = 0$ | 30 | 1.0E-02 |
| Salomon Problem | $f_2(x) = 1 - \cos(2\pi \sqrt{\sum_{i=1}^{D} x_i^2}) + 0.1(\sqrt{\sum_{i=1}^{D} x_i^2})$ | [-100, 100] | $f(0) = 0$ | 30 | 1.0E-01 |
| Colville function | $f_3(x) = 100(x_2 - x_1^2)^2 + (1 - x_1)^2 + 90(x_4 - x_3^2)^2 + (1 - x_3)^2 + 10.1[(x_2 - 1)^2 + (x_4 - 1)^2] + 19.8(x_2 - 1)(x_4 - 1)$ | [-10, 10] | $f(1) = 0$ | 4 | 1.0E-05 |
| Braninss Function | $f_4(x) = a(x_2 - bx_1^2 + cx_1 - d)^2 + e(1-f)\cos x_1 + e$ | $x_1 \in [-5, 10]$, $x_2 \in [0, 15]$ | $f(-\pi, 12.275) = 0.3979$ | 2 | 1.0E-05 |
| Kowalik function | $f_5(x) = \sum_{i=1}^{11} (a_i - \frac{x_1(b_i^2 + b_i x_2)}{b_i^2 + b_i x_3 + x_4})^2$ | [-5, 5] | $f(0.1928, 0.1908, 0.1231, 0.1357) = 3.07E\text{-}04$ | 4 | 1.0E-05 |
| Shifted Rosenbrock | $f_6(x) = \sum_{i=1}^{D-1}(100(z_i^2 - z_{i+1})^2 + (z_i - 1)^2 + f_{bias}$, $z = x - o + 1, x = [x_1, x_2, ... x_D], o = [o_1, o_2, ...... o_D]$ | [-100, 100] | $f(o) = f_{bias} = 390$ | 10 | 1.0E-01 |
| Six-hump camel back | $f_7(x) = (4 - 2.1 x_1^2 + \frac{1}{3} x_1^4) x_1^2 + x_1 x_2 + (-4 + 4 x_2^2) x_2^2$ | [-5, 5] | $f(-0.0898, 0.7126) = -1.0316$ | 2 | 1.0E-05 |
| Easom's function | $f_8(x) = -\cos x_1 \cos x_2 e^{(-(x_1 - \pi)^2 - (x_2 - \pi)^2)}$ | [-10, 10] | $f(\pi, \pi) = -1$ | 2 | 1.0E-13 |
| Hosaki Problem | $f_9(x) = (1 - 8x_1 + 7 x_1^2 - \frac{7}{3} x_1^3 + \frac{1}{4} x_1^4) x_2^2 \exp(-x_2)$ | $x_1 \in [0, 5]$, $x_2 \in [0, 6]$ | -2.3458 | 2 | 1.0E-06 |
| McCormick | $f_{10}(x) = \sin(x_1 + x_2) + (x_1 - x_2)^2 - \frac{3}{2} x_1 + \frac{5}{2} x_2 + 1$ | $-1.5 \le x_1 \le 4$, $-3 \le x_2 \le 3$ | $f(-0.547, -1.547) = -1.9133$ | 30 | 1.0E-04 |
| Meyer and Roth Problem | $f_{11}(x) = \sum_{i=1}^{5} (\frac{x_1 x_3 t_i}{1 + x_1 t_i + x_2 v_i} - y_i)^2$ | [-10, 10] | $f(3.13, 15.16, 0.78) = 0.4E\text{-}04$ | 3 | 1.0E-03 |
| Shubert | $f_{12}(x) = -\sum_{i=1}^{5} i \cos((i+1)x_1 + 1) \sum_{i=1}^{5} i \cos((i+1)x_2 + 1)$ | [-10, 10] | $f(7.0835, 4.8580) = -186.7309$ | 2 | 1.0E-05 |

**Table 2. Comparison of the results of test problems**

| Test Problem | Algorithm | MFV | SD | ME | AFE | SR |
|---|---|---|---|---|---|---|
| $f_1$ | ABC | 9.92E+01 | 1.49E+01 | 9.92E+01 | 100020 | 0 |
| | RMABC | 9.62E+01 | 1.78E+01 | 9.62E+01 | 200000 | 0 |
| | MeABC | 2.01E-02 | 7.87E-03 | 2.01E-02 | 99739.44 | 5 |
| | EnABC | 9.67E-03 | 3.78E-04 | 9.67E-03 | 115431.8 | 100 |
| | IoABC | 1.02E-02 | 1.32E-03 | 1.02E-02 | 88594.8 | 79 |
| $f_2$ | ABC | 1.67E+00 | 2.30E-01 | 1.67E+00 | 100020.1 | 0 |
| | RMABC | 9.36E-01 | 3.32E-02 | 9.36E-01 | 87168.16 | 97 |
| | MeABC | 9.18E-01 | 3.47E-02 | 9.18E-01 | 20766.84 | 100 |
| | EnABC | 9.21E-01 | 3.05E-02 | 9.21E-01 | 29277.91 | 100 |
| | IoABC | 9.14E-01 | 4.05E-02 | 9.14E-01 | 18276.72 | 100 |
| $f_3$ | ABC | 1.93E-01 | 1.54E-01 | 1.93E-01 | 99123.47 | 1 |
| | RMABC | 1.80E-02 | 1.62E-02 | 1.80E-02 | 159293.8 | 41 |
| | MeABC | 7.53E-03 | 3.29E-03 | 7.53E-03 | 39822.68 | 90 |
| | EnABC | 1.59E-02 | 1.72E-02 | 1.59E-02 | 112400.9 | 60 |
| | IoABC | 7.64E-03 | 2.52E-03 | 7.64E-03 | 36752.5 | 98 |
| $f_4$ | ABC | 3.98E-01 | 6.26E-06 | 5.74E-06 | 13070.99 | 89 |
| | RMABC | 3.98E-01 | 6.33E-06 | 5.49E-06 | 19134.85 | 91 |
| | MeABC | 3.98E-01 | 6.84E-06 | 6.30E-06 | 14998.47 | 86 |
| | EnABC | 3.98E-01 | 6.98E-06 | 6.29E-06 | 27786.34 | 87 |
| | IoABC | 3.98E-01 | 6.66E-06 | 5.97E-06 | 12110.51 | 89 |
| $f_5$ | ABC | 4.72E-04 | 7.13E-05 | 1.65E-04 | 87107.16 | 32 |
| | RMABC | 3.96E-04 | 3.45E-05 | 8.86E-05 | 91857.83 | 89 |
| | MeABC | 3.98E-01 | 6.84E-06 | 6.30E-06 | 14998.47 | 86 |
| | EnABC | 3.91E-04 | 2.40E-05 | 8.39E-05 | 73715.78 | 95 |
| | IoABC | 4.16E-04 | 6.82E-05 | 1.08E-04 | 58993.74 | 84 |
| $f_6$ | ABC | 3.95E+02 | 5.11E+00 | 5.03E+00 | 95138.97 | 10 |
| | RMABC | 3.90E+02 | 3.82E-02 | 9.02E-02 | 101917.5 | 91 |
| | MeABC | 3.91E+02 | 1.10E+00 | 6.52E-01 | 77067.59 | 40 |
| | EnABC | 3.92E+02 | 3.24E+00 | 2.17E+00 | 171792.1 | 21 |
| | IoABC | 3.91E+02 | 2.73E+00 | 1.10E+00 | 80512.48 | 40 |
| $f_7$ | ABC | 3.00E+00 | 3.43E-05 | 5.15E-06 | 78504.88 | 34 |
| | RMABC | -1.03E+00 | 7.52E-04 | 4.03E+00 | 200027.9 | 0 |
| | MeABC | 3.00E+00 | 4.13E-15 | 4.96E-15 | 5770.3 | 100 |
| | EnABC | 3.00E+00 | 9.28E-09 | 9.59E-10 | 106115.1 | 73 |
| | IoABC | 3.00E+00 | 4.27E-15 | 4.66E-15 | 7913.22 | 100 |
| $f_8$ | ABC | -1.03E+00 | 3.33E-03 | 4.06E-03 | 100040.6 | 0 |
| | RMABC | -9.85E-01 | 1.37E-02 | 4.68E-02 | 200026 | 0 |
| | MeABC | -1.03E+00 | 1.35E-05 | 1.53E-05 | 55900.32 | 49 |
| | EnABC | -1.03E+00 | 1.37E-05 | 1.68E-05 | 121099.5 | 43 |
| | IoABC | -1.03E+00 | 1.37E-05 | 1.75E-05 | 58389.18 | 44 |
| $f_9$ | ABC | -2.47E+04 | 7.61E+01 | 6.14E+01 | 100041.6 | 0 |
| | RMABC | -5.99E+11 | 1.43E+11 | 5.99E+11 | 200025 | 0 |
| | MeABC | -2.48E+04 | 1.66E-01 | 6.05E-01 | 86608.66 | 23 |
| | EnABC | -2.48E+04 | 1.28E-01 | 6.00E-01 | 181448.8 | 16 |
| | IoABC | -2.48E+04 | 3.18E-02 | 5.10E-01 | 73568.61 | 49 |
| $f_{10}$ | ABC | -2.31E+00 | 3.25E-02 | 3.39E-02 | 100033.1 | 0 |
| | RMABC | 9.35E-01 | 4.44E-16 | 3.28E+00 | 200021.2 | 0 |
| | MeABC | -2.35E+00 | 1.69E-05 | 1.01E-05 | 50091.38 | 75 |
| | EnABC | -2.35E+00 | 7.69E-06 | 6.11E-06 | 75705.21 | 88 |
| | IoABC | -2.35E+00 | 7.16E-06 | 6.71E-06 | 35178.05 | 84 |

**Table 2. Comparison of the results of test problems (cont.)**

| Test Problem | Algorithm | MFV | SD | ME | AFE | SR |
|---|---|---|---|---|---|---|
| $f_{11}$ | ABC | -1.88E+00 | 3.05E-02 | 3.28E-02 | 100035.8 | 0 |
| | RMABC | 1.50E-02 | 1.73E-18 | 1.93E+00 | 200022.7 | 0 |
| | MeABC | -1.91E+00 | 1.16E-05 | 9.11E-05 | 41514.96 | 86 |
| | EnABC | -1.91E+00 | 4.56E-05 | 1.20E-04 | 137127 | 48 |
| | IoABC | -1.91E+00 | 7.00E-06 | 8.74E-05 | 18802.85 | 100 |
| $f_{12}$ | ABC | 1.91E-03 | 6.39E-06 | 1.95E-03 | 31617.35 | 91 |
| | RMABC | -1.79E+02 | 7.07E+00 | 1.79E+02 | 200023 | 0 |
| | MeABC | 1.91E-03 | 2.81E-06 | 1.95E-03 | 4658.79 | 100 |
| | EnABC | 1.91E-03 | 2.93E-06 | 1.95E-03 | 12426.8 | 100 |
| | IoABC | 1.91E-03 | 2.92E-06 | 1.95E-03 | 4355.93 | 100 |
| $f_{13}$ | ABC | 2.65E+00 | 1.14E-02 | 2.30E-02 | 96378.65 | 8 |
| | RMABC | 2.64E+00 | 1.06E-02 | 1.11E-02 | 185673.8 | 13 |
| | MeABC | 2.63E+00 | 9.37E-03 | 8.04E-03 | 93536.88 | 15 |
| | EnABC | 2.63E+00 | 9.07E-03 | 7.37E-03 | 177131.7 | 25 |
| | IoABC | 2.63E+00 | 3.58E-03 | 3.05E-03 | 71783.43 | 51 |
| $f_{14}$ | ABC | 3.69E-13 | 5.69E-13 | 3.69E-13 | 100020 | 0 |
| | RMABC | -1.61E+38 | 8.83E+37 | 1.61E+38 | 200015.1 | 0 |
| | MeABC | 8.99E-16 | 1.02E-16 | 8.99E-16 | 63060.48 | 100 |
| | EnABC | 9.13E-16 | 6.64E-17 | 9.13E-16 | 153553.9 | 100 |
| | IoABC | 9.02E-16 | 9.14E-17 | 9.02E-16 | 52811.1 | 100 |

The search boundaries for the variables are

$1.125 \leq x_1 \leq 12.5$, $0.625 \leq x_2 \leq 12.5$,

$1.0*10^{-8} \leq x_3 \leq 240$ and $1.0*10^{-8} \leq x_4 \leq 240$.

The best ever identified global optimum solution is $f(1.125, 0.625, 55.8592, 57.7315) = 7197.729$ [41]. The tolerable error for considered problem is $1.0E-05$.

## 5.2 Experimental Setup

To prove the efficiency of IoABC, it is compared with original ABC algorithm [6], Randomized Memetic ABC (RMABC) algorithm [13], Memetic search in ABC (MeABC) algorithm [34] and Enhanced Local Search in Artificial Bee Colony (EnABC) Algorithm [12] over well thought-out fourteen problems, following experimental setting is adopted:

- The size of colony= Population size SN =60
- Number of Employed bee = Number of Onlooker bee =SN/2
- The maximum number of cycles for foraging MCN =200000
- Number of repetition of experiment =Runtime =100
- Limit =1500 ,A food source which could not be improved through "limit" trial is abandoned by its employed bee
- The mean function values (MFV), standard deviation (SD), mean error (ME), average function evaluation (AFE) and success rate (SR) of considered problem have been recorded.
- Experimental setting for ABC, RMABC, MeABC and EnABC are same as IoABC.

## 5.3 Result Comparison

Mathematical results of IoABC with experimental setting as per subsection 5.2 are discussed in Table 2. Table 2 show the connection of results based on mean function value (MFV), standard deviation (SD), mean error (ME), average function evaluations (AFE) and success rate (SR). Table 2 shows that a good number of the times IoABC outperforms in terms of efficiency (with less number of function evaluations) and reliability as compare to other considered algorithms. The proposed algorithm all the time improves AFE and most of the time it also improve SD and ME. It is due to randomness introduced during fitness calculation and probability calculation. Table 3 contains summary of table 2 outcomes. In Table 3, '+' indicates that the IoABC is better than the considered algorithms and '-' indicates that the algorithm is not better or the difference is very small. The last row of Table 3, establishes the superiority of IoABC over RMABC, EnABC, MeABC and ABC.

## 6. CONCLUSION

This paper, modify the onlooker bee phase in original ABC by introducing modified GSS process. Newly introduced strategy added in onlooker bee phase. Proposed algorithm modifies search range of GSS process and solution update equation in order to balance intensification and diversification of local search space. Further, the modified strategy is applied to solve 12 well-known standard benchmark functions and two real world problems. With the help of experiments over test problems and real world problems, it is shown that the insertion of the proposed strategy in the original ABC algorithm improves the steadfastness, efficiency and accuracy as compare to their original version. Table 2 and 3 show that the proposed IoABC is able to solve almost all the considered problems with fewer efforts. Numerical results show that the improved algorithm is superior to original ABC algorithm and its recent variants. Proposed algorithm has the ability to get out of a local minimum and has higher rate of convergence. It can be resourcefully applied for separable, multivariable, multimodal function optimization. The proposed strategy also improves results for two real world problems: compression spring and pressure vessel design. When applied to solve compression spring it improves results by 50% (in term of success rate and average function evaluation).

Table 3. Summary of table 2 outcome

| Test Problem | IoABC vs. ABC | IoABC vs. RMABC | IoABC vs. MeABC | IoABC vs. EnABC |
|---|---|---|---|---|
| $f_1$ | + | + | + | - |
| $f_2$ | + | + | + | + |
| $f_3$ | + | + | + | + |
| $f_4$ | + | - | + | + |
| $f_5$ | + | - | - | - |
| $f_6$ | + | - | - | + |
| $f_7$ | + | + | - | + |
| $f_8$ | + | + | - | + |
| $f_9$ | + | + | + | + |
| $f_{10}$ | + | + | + | - |
| $f_{11}$ | + | + | + | + |
| $f_{12}$ | + | + | + | - |
| $f_{13}$ | + | + | + | + |
| $f_{14}$ | + | + | + | + |
| Total Number of + Sign | 14 | 11 | 10 | 10 |

## 7. REFERENCES

[1] M Dorigo, G Di Caro (1999) Ant colony optimization: a new metaheuristic. In: Evolutionary computation, 1999. CEC 99. Proceedings of the 1999 congress on, 2. IEEE

[2] J Kennedy, R Eberhart (1995) Particle swarm optimization. In: Neural networks, 1995. Proceedings., IEEE international conference on, 4, pp 1942–1948. IEEE

[3] KV Price, RM Storn, JA Lampinen (2005) Differential evolution: a practical approach to global optimization. Springer, Berlin

[4] J Vesterstrom, R Thomsen (2004) A comparative study of differential evolution, particle swarm optimization, and evolutionary algorithms on numerical benchmark problems. In: Evolutionary computation, 2004. CEC2004. Congress on, 2, pp 1980–1987. IEEE

[5] KM Passino (2002) Biomimicry of bacterial foraging for distributed optimization and control. IEEE Control SystMag 22(3):52–67

[6] D Karaboga (2005) An idea based on honey bee swarm for numerical optimization. Techn. Rep. TR06, Erciyes University Press, Erciyes

[7] D Karaboga, B Akay (2009) A comparative study of artificial bee colony algorithm. Appl Math Comput 214(1):108–132

[8] G Zhu, S Kwong (2010) Gbest-guided artificial bee colony algorithm for numerical function optimization. Appl Math Comput 217(7):3166–3173

[9] F Kang, J Li, Z Ma, H Li (2011) Artificial bee colony algorithm with local search for numerical optimization. J Softw 6(3):490–497

[10] R Hooke, TA Jeeves (1961) "Direct search" solution of numerical and statistical problems. J ACM (JACM) 8(2):212–229

[11] S Kumar, VK Sharma and R Kumari (2013) A Novel Hybrid Crossover based Artificial Bee Colony Algorithm for Optimization Problem, International Journal of Computer Application 82(8):18-25

[12] S Kumar and VK Sharma (2014), Enhanced Local Search in Artificial Bee Colony Algorithm. International Journal of Emerging Technologies in Computational and Applied Sciences (IJETCAS). In Print.

[13] S Kumar, VK Sharma and R Kumari (2014) Randomized Memetic Artificial Bee Colony Algorithm. International Journal of Emerging Trends & Technology in Computer Science (IJETTCS).In Print.

[14] S Kumar and VK Sharma (2014) Modified Artificial Bee Colony Algorithm. International Journal of Information, Communication and Computing Technology. In Print.

[15] S Kumar, VK Sharma, A Kumar and H Sharma (2014) Fitness Based Position Update in Artificial Bee Colony Algorithm. Unpublished.

[16] DE Goldberg (1989) Genetic algorithms in search, optimization, and machine learning.

[17] R Storn and K Price (1997) Differential evolution-a simple and efficient adaptive scheme for global optimization over continuous spaces. Journal of Global Optimization, 11:341–359.

[18] PA Moscato, "On evolution, search, optimization, genetic algorithms and martial arts: Towards memetic algorithms." Tech. Rep. Caltech Concurrent Computation Program, Report. 826, California Inst. of Tech., Pasadena, California, USA (1989).

[19] E Burke, J Newall, R Weare.: A memetic algorithm for university exam timetabling. In: E. Burke, P. Ross (eds.) The Practice and Theory of Automated Timetabling, Lecture Notes in Computer Science, vol. 1153, pp. 241–250. Springer Verlag (1996).

[20] R Carr, W Hart, N Krasnogor, E Burke, J Hirst, J Smith, "Alignment of protein structures with a memetic

evolutionary algorithm," In: Proceedings of the Genetic and Evolutionary Computation Conference. Morgan Kaufman (2002).

[21] R Cheng, M Gen, "Parallel machine scheduling problems using memetic algorithms." Computers & Industrial Engineering 33(3–4), 761–764 (1997).

[22] C Fleurent, J Ferland, "Genetic and hybrid algorithms for graph coloring." Annals of Operations Research 63, 437–461 (1997).

[23] G Gutin, D Karapetyan, N Krasnogor, "Memetic algorithm for the generalized asymmetric traveling salesman problem." In: M. Pavone, G. Nicosia, D. Pelta, N. Krasnogor (eds.) Proceedings of the 2007Workshop On Nature Inspired Cooperative Strategies for Optimisation. Lecture Notes in Computer Science (LNCS), vol. to appear. Springer (2007)

[24] L He, N Mort, "Hybrid genetic algorithms for telecomunications network back-up routeing." BT Technology Journal 18(4) (2000)

[25] C Reeves, "Hybrid genetic algorithms for bin-packing and related problems." Annals of Operations Research 63, 371–396 (1996)

[26] M Tang, X Yao, "A memetic algorithm for VLSI floorplanning." Systems, Man, and Cybernetics, Part B, IEEE Transactions on 37(1), 62–69 (2007). DOI 10.1109/TSMCB.2006.883268.

[27] WE Hart, "Adaptive Global Optimization with Local Search." Ph.D. Thesis, University of California, San Diego (1994)

[28] GM Morris, DS Goodsell, RS Halliday, R Huey, WE Hart, RK Belew, AJ Olson,: "Automated docking using a lamarkian genetic algorithm and an empirical binding free energy function." J Comp Chem 14, 1639–1662 (1998)

[29] H Wang, D Wang, and S Yang. "A memetic algorithm with adaptive hill climbing strategy for dynamic optimization problems." Soft Computing 13.8-9 (2009): 763-780.

[30] D Liu, KC Tan, CK Goh, WK Ho, "A multi-objective memetic algorithm based on particle swarm optimization." Systems, Man, and Cybernetics, Part B, IEEE Transactions on 37(1), 42–50 (2007).

[31] Y Wang, JK Hao, F Glover, Z Lü, "A tabu search based memetic algorithm for the maximum diversity problem." Engineering Applications of Artificial Intelligence 27 (2014): 103-114.

[32] X Xue, Y Wang, and A Ren. "Optimizing ontology alignment through Memetic Algorithm based on Partial Reference Alignment." Expert Systems with Applications 41.7 (2014): 3213-3222.

[33] O Chertov and D Tavrov. "Memetic Algorithm for Solving the Task of Providing Group Anonymity." Advance Trends in Soft Computing. Springer International Publishing, 2014. 281-292.

[34] JC Bansal, H Sharma, KV Arya and A Nagar, "Memetic search in artificial bee colony algorithm." Soft Computing (2013): 1-18.

[35] I Fister, I Fister Jr, J Bres, V Zumer. "Memetic artificial bee colony algorithm for large-scale global optimization." Evolutionary Computation (CEC), 2012 IEEE Congress on. IEEE, 2012.

[36] J Kiefer (1953) Sequential minimax search for a maximum. In: Proceedings of American Mathematical Society, vol. 4, pp 502–506.

[37] H Sharma, JC Bansal, and KV Arya. "Opposition based lévy flight artificial bee colony." *Memetic Computing* (2012): 1-15.

[38] MM Ali, C Khompatraporn, and ZB Zabinsky. "A numerical evaluation of several stochastic algorithms on selected continuous global optimization test problems." Journal of Global Optimization, 31(4):635–672, 2005.

[39] PN Suganthan, N Hansen, JJ Liang, K Deb, YP Chen, A. Auger, and S. Tiwari. "Problem definitions and evaluation criteria for the CEC 2005 special session on real-parameter optimization." In CEC 2005, 2005.